\documentclass[10pt,twocolumn,letterpaper]{article}
\usepackage[pagenumbers]{cvpr}
\usepackage{mathtools,colortbl,multirow,enumitem,microtype,placeins}

\usepackage{amsmath,amsfonts,bm}

\def\eqref#1{equation~\ref{#1}}

\def\1{\bm{1}}

\DeclareMathAlphabet{\mathsfit}{\encodingdefault}{\sfdefault}{m}{sl}
\SetMathAlphabet{\mathsfit}{bold}{\encodingdefault}{\sfdefault}{bx}{n}

\graphicspath{{images/}}
\ifdefined\pdfminorversion\pdfminorversion=7\fi

\providecommand{\Description}[1]{}
\newcommand{\tablestyle}[2]{%
  \setlength{\tabcolsep}{#1}%
  \renewcommand{\arraystretch}{#2}%
  \centering\footnotesize}
\definecolor{graycolor}{gray}{.9}

\newcommand{\best}[1]{\textbf{#1}}
\newcommand{\bestmath}[1]{\bm{#1}}

\definecolor{cvprblue}{rgb}{0.21,0.49,0.74}
\usepackage[pagebackref,breaklinks,colorlinks,allcolors=cvprblue,hypertexnames=false]{hyperref}
\def\confName{CVPR}
\def\confYear{2027}
\title{Octree-based Video Representation}
\author{Rungui Zhou \quad Chuanzhi Zhou \quad Yuk-Kit Hou \quad Peng-Shuai Wang\\
Peking University}
\hypersetup{pdftitle={Octree-based Video Representation},
pdfauthor={Rungui Zhou, Chuanzhi Zhou, Yuk-Kit Hou, Peng-Shuai Wang},
pdfsubject={Preprint}}
\begin{document}
\maketitle
\begin{figure*}[t]
  \centering
  \includegraphics[width=\textwidth]{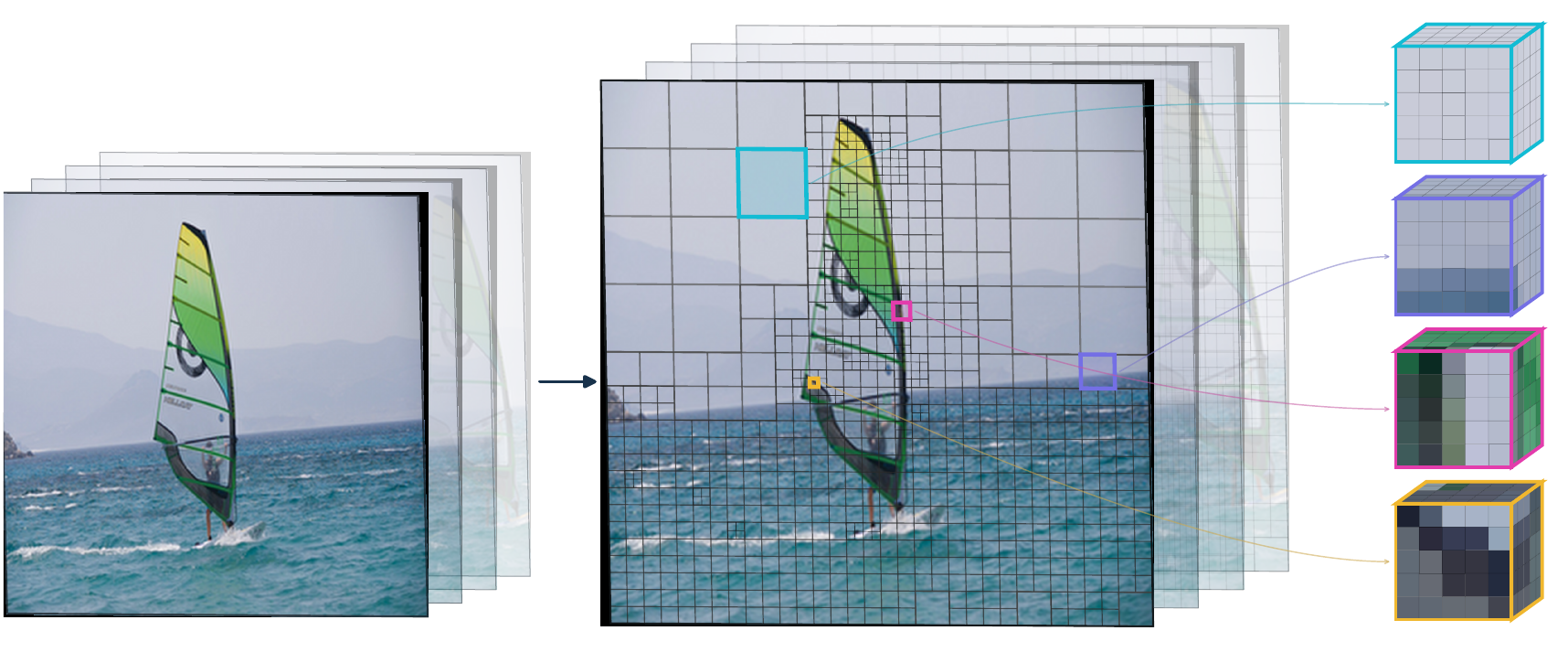}
  \caption{Content-adaptive video representation with OctVideo.
  A video volume is partitioned into large cells in smooth regions and finer
  cells where local approximation error is high. Colored boxes mark the cells
  enlarged on the right, illustrating different spatio-temporal scales.}
  \label{fig:teaser}
\end{figure*}

\begin{abstract}
Representing videos on uniform spatio-temporal grids produces large token sequences and substantial computational costs. Yet visual complexity is unevenly distributed: smooth regions can be represented compactly, while detailed regions require finer resolution. We introduce OctVideo, an
octree-based video representation that adapts cell size to local approximation
error. Each leaf stores local RGB values and spatio-temporal gradients,
supporting analytic reconstruction of the video. A lightweight learned
residual enriches these descriptors for neural processing. For reconstruction,
a Conv1D VAE maps the serialized cells to a
regular latent grid and selectively refines details during decoding. Our VAE
achieves 36.12 dB PSNR with 38.2M parameters and 188.9 GFLOPs per clip on
Kinetics-400 (K400). It also generalizes zero-shot to the high-resolution
Densely Annotated VIdeo Segmentation (DAVIS) 2016 dataset with reconstruction
quality comparable to the best evaluated models. On both datasets, it
requires the fewest model FLOPs and achieves the fastest reconstruction
among the evaluated models. OctVideo also supports video
understanding, achieving competitive recognition performance with few input
tokens when trained from scratch. By exploiting the redundancy already
present in video signals and efficiently processing sparse structures,
OctVideo provides an efficient representation for video.
\end{abstract}

\section{Introduction} \label{sec:intro}

Video content varies in complexity across space and time, yet video models commonly process it on uniform grids during training and inference.
A slowly varying background can occupy most of a clip, while fine detail is concentrated around object boundaries, textured surfaces, and transient events.
Fixed-size tokens assign the same sampling density to these regions despite their different representational needs.
An efficient video representation should allocate finer cells where the signal requires them, preserve a local approximation throughout the video, and expose enough structure for neural networks to process it efficiently.

Existing approaches address different aspects of this problem.
Classical and learned codecs remove redundancy primarily for storage and transmission~\citep{Wiegand2003AVC,Lu2019DVC}.
Video VAEs learn compact latent grids~\citep{wan21,hunyuanvae}, with efficient variants using wavelets and compressed sensing to reduce computation~\citep{WF-WAE,leanvae}.
These architectures improve processing efficiency on regular grids with fixed input cell sizes.
For recognition, token pruning and merging reduce feature computation by selecting or combining tokens~\citep{hao2025lite,bolya2023tome}.
Our goal is a compact input representation that combines content-adaptive allocation, direct reconstruction, and efficient neural processing.

Classical graphics methods have been using adaptive and hierarchical data structures to represent smooth regions coarsely and complex regions finely~\citep{Frisken2000,Ohtake2003}.
Learned octree and quadtree methods bring related principles to shapes and images~\citep{Wang2018a,Wang2022,Wang2023,Ronen2023}.
Applying this idea to video poses two linked challenges: retaining local RGB variation within cells that span both space and time, and translating an irregular partition into computational savings in neural models.

In this paper, we introduce \textbf{OctVideo}, an octree-based representation that couples adaptive partitioning with local signal approximation.
Each cell stores \emph{a fixed grid of RGB values and spatio-temporal gradients}, defining local first-order approximations.
The approximation error determines whether the cell is subdivided: smooth regions remain coarse, while regions with larger errors receive finer cells.
The same local approximation determines both the stored descriptors and the refinement criterion.
The octree leaves cover the entire video volume, so every location supports analytic RGB reconstruction from its covering leaf.
For neural processing, a lightweight convolutional branch supplies learned residual features that enrich each leaf while preserving the error-guided partition.

To turn this compact representation into efficient neural computation, we design a VAE around the octree structure.
The encoder serializes mixed-depth leaves in Morton order and processes them with regular 1D convolutional kernels.
Cell bounds then determine how encoded features are copied or pooled onto a regular latent grid.
A hierarchical decoder learns from octree-derived refinement targets to predict which latent cells require finer processing.
This design carries adaptive allocation from the input representation into neural encoding and decoding.
The same representation supports recognition through patch-wise attention and sibling pooling along the octree hierarchy.

On Kinetics-400 (K400)~\citep{Carreira_2017_CVPR}, the VAE takes an average of 3,657.66 adaptive cells as input, $17.9\times$ fewer than the 65,536 patches in the regular finest-level partition.
Analytic reconstruction alone achieves 34.72 dB PSNR, demonstrating the fidelity of the representation before neural decoding.
The learned VAE achieves 36.12 dB with 188.9 GFLOPs.
Under the reported evaluation protocol, it decodes nearly two orders of magnitude faster than IV-VAE, the highest-PSNR baseline, with a 0.69 dB lower PSNR.
Compared with the lightweight LeanVAE, OctVideo improves PSNR by 0.66 dB while using over 40\% fewer model FLOPs and decoding $2.3\times$ faster.
Zero-shot transfer to high-resolution DAVIS 2016 videos~\citep{perazzi2016davis} yields 32.98 dB PSNR with a VAE runtime of 0.54 seconds per clip using spatial tiling.
A further experiment on UCF101 demonstrates that the same K400-trained VAE supports class-conditional video generation with flow matching.
For recognition, OctVideo trained from scratch achieves 80.1\% K400 Top-1 accuracy with approximately 4,200 input tokens per view, within 0.1 percentage points of MViTv1-B using about $12\times$ fewer input tokens under the respective evaluation protocols.
In summary, our contributions are:
\begin{itemize}[leftmargin=*,topsep=3pt,itemsep=2pt,parsep=0pt,partopsep=0pt]
  \item An adaptive video representation in which first-order RGB approximation guides octree refinement and preserves a directly reconstructible signal, with learned residual features for neural processing.
  \item An efficient VAE that connects mixed-depth cells to a regular latent grid and uses octree-supervised selective refinement to concentrate decoding computation where finer detail is needed.
  \item Experiments demonstrate substantial reductions in tokens, VAE compute,
  and decoding time while retaining reconstruction quality and recognition accuracy.
\end{itemize}
Our code will be made publicly available.

\section{Related Work} \label{sec:related}

Classical video codecs remove statistical redundancy through motion-compensated
prediction, block transforms, quantization, and entropy coding. Standards such
as H.264/AVC and HEVC, progressively enrich the partition and prediction choices used for rate--distortion
optimization~\citep{Wiegand2003AVC,Sullivan2012HEVC}. Learned codecs replace
parts of this hand-designed pipeline with neural motion estimation and
analysis--synthesis transforms~\citep{Lu2019DVC,Agustsson2020SSF}.
Compressed-domain recognition methods such as CoViAR learn from I-frames,
motion vectors, and prediction residuals~\citep{wu2018coviar}.
OctVideo instead constructs an explicit
spatio-temporal signal representation whose cells can be processed directly and
also decoded locally.

Hierarchical local approximation underlies adaptive distance fields and
partition-of-unity surfaces~\citep{Frisken2000,Ohtake2003}. Adaptive O-CNN,
dual octree graph networks, and OctFormer extend octree structures to learned
shape representations and point-cloud processing~\citep{Wang2018a,Wang2022,Wang2023},
while Quadformer uses saliency-guided quadtrees for mixed-resolution image
tokenization~\citep{Ronen2023}. OctVideo applies adaptive local approximation
to a dense RGB field over space and time. Its value-gradient descriptors
support analytic reconstruction and provide the error criterion for
refinement, while the resulting tokens and hierarchy support learned
recognition and reconstruction.

VAEs learn continuous latent representations~\citep{Kingma2013}, providing
a compact space for latent diffusion~\citep{Rombach2022}.
VQ-VAE introduced discrete learned codes, which VideoGPT extended to video
with a spatio-temporal encoder~\citep{Van2017,Yan2021VideoGPT}. Modern video
generation systems more often use continuous latent grids: Wan, HunyuanVideo,
CogVideoX, CV-VAE, and IV-VAE employ spatio-temporal encoders and
decoders~\citep{wan21,hunyuanvae,cogvideox,cvvae,IV-VAE}. WF-VAE and LeanVAE
reduce their cost through wavelet processing, with LeanVAE additionally using
compressed sensing~\citep{WF-WAE,leanvae}. More recent work further improves the latent representation itself: PyraTok learns language-aligned multi-scale codes, PV-VAE adds future prediction, and KATok performs content-dependent token dropping~\citep{Susladkar2026PyraTok,Zhao2026PVVAE,Lee2026KATok}. 
These learned latents rely on neural decoders for reconstruction. OctVideo's
input cells instead carry local RGB approximations that can be evaluated
analytically, while preserving full spatio-temporal coverage.

Efficient video architectures vary temporal sampling and feature
resolution~\citep{feichtenhofer2019slowfast,fan2021multiscale,li2022mvitv2}.
Token reduction methods instead prune unimportant tokens, as in DynamicViT for
images, select subsets for video recognition, as in LITE, or merge similar
features, as in ToMe~\citep{rao2021dynamicvit,hao2025lite,bolya2023tome}.
OneVision-Encoder uses codec information to select informative RGB
regions~\citep{onevision}. These approaches reduce feature computation but
do not explicitly preserve a locally decodable RGB approximation for every
region. OctVideo instead coarsens the signal itself: smooth regions use larger
cells storing RGB values and gradients, while high-error regions are refined.
 Every location therefore remains represented, enabling the
same compact input to support both understanding and reconstruction.

\section{Method} \label{sec:method}

\begin{figure*}[t]
  \centering
  \includegraphics[width=\textwidth]{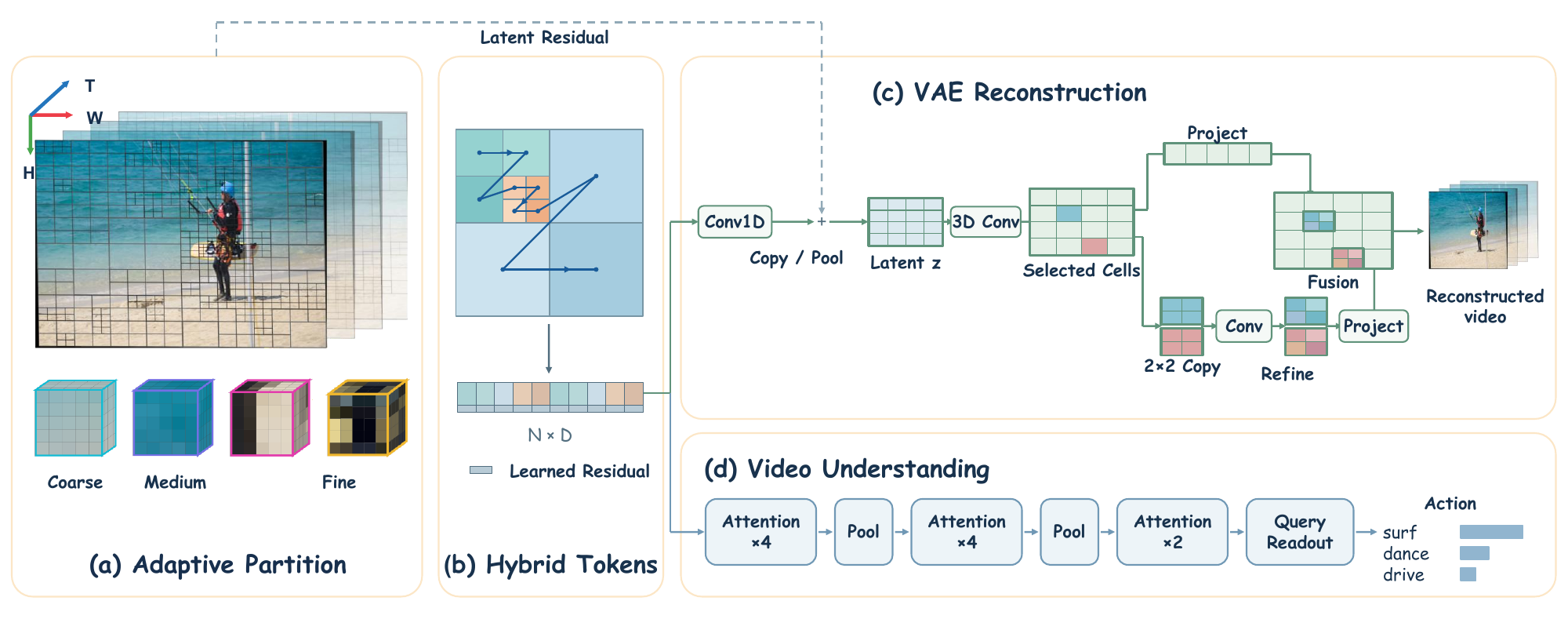}
  \caption{Overview of OctVideo. (a) Error-guided octree partitioning produces
  cells of different sizes, shown as equal-sized tokens for illustration.
  (b) Analytic descriptors and learned residual features form a Morton-ordered
  token sequence. (c) Conv1D encoding and copy/pool mapping produce a regular
  latent grid, supplemented by the dense latent residual shown as a dashed
  connection. The decoder refines selected cells and fuses them with coarse
  regions to reconstruct the video; 2D grids illustrate the 3D operations.
  (d) Hierarchical attention and sibling pooling support action recognition.}
  \Description{Four panels show adaptive octree partitioning, hybrid token
  serialization, VAE reconstruction with selective refinement and fusion,
  and attention-based action recognition. A top dashed connection adds a
  dense residual to the latent representation.}
  \label{fig:method-overview}
\end{figure*}

OctVideo allocates cells according to local approximation error, giving smooth
regions a compact description and preserving finer structure where needed.
We first describe the shared representation and efficient VAE for
reconstruction, followed by the hierarchical classifier used for complementary
recognition evaluation
(\cref{fig:method-overview}).

\subsection{OctVideo Representation}

Let $V:\Omega\rightarrow\mathbb{R}^{3}$ be an ImageNet-standardized RGB video with coordinates
$\mathbf{x}=(t,h,w)$. Each octree cell $c$ covers a region $\Omega_c$
partitioned into a fixed $2\times4\times4$ grid of subregions
$\Omega_{c,q}$, indexed by $q=1,\ldots,32$. The grid dimensions are fixed,
while its video coverage varies with the cell's depth. Within each subregion,
\begin{equation}
  \hat V_{c,q}(\mathbf{x})
  =\mathbf{a}_{c,q}+\mathbf{G}_{c,q}\mathbf{u}_{c,q}(\mathbf{x}),
  \qquad \mathbf{x}\in\Omega_{c,q}.
\end{equation}
Here $\mathbf{u}_{c,q}$ denotes centered local coordinates normalized to
$[-\frac12,\frac12]$ along each axis (zero for a singleton axis).
The RGB value $\mathbf{a}_{c,q}\in\mathbb{R}^{3}$ and gradients
$\mathbf{G}_{c,q}\in\mathbb{R}^{3\times3}$ with respect to these coordinates
form a 12-dimensional descriptor. For subregions containing multiple voxels,
these coefficients are fitted by local least squares:
\begin{equation}
  (\mathbf{a}_{c,q},\mathbf{G}_{c,q})
  =\arg\min_{\mathbf{a},\mathbf{G}}
  \sum_{\mathbf{x}\in\Omega_{c,q}}
  \|V(\mathbf{x})-\mathbf{a}-\mathbf{G}\mathbf{u}_{c,q}(\mathbf{x})\|_2^2.
  \label{eq:local-ls}
\end{equation}

We define the maximum local approximation error as
\begin{equation}
  e(c)=\max_{q}\max_{\mathbf{x}\in\Omega_{c,q}}
  \|V(\mathbf{x})-\hat V_{c,q}(\mathbf{x})\|_\infty.
\end{equation}
Starting from a coarse partition, we split a cell into eight children when
$d_c<D$ and $e(c)>\tau_{d_c}$, where $D$ is the finest depth.
Both tasks use $(\tau_3,\tau_4,\tau_5)=(0.7,0.8,1.0)$ in this standardized
space, with refinement stopping at depth 6.
The tree construction procedure is detailed in
\cref{sec:appendix-construction}.

Each retained leaf becomes one token
$\mathbf{f}_c=\operatorname{Flatten}([\mathbf{a}_{c,q},\mathbf{G}_{c,q}]_{q=1}^{32})
\in\mathbb{R}^{384}$, regardless of its coverage. Together with its center
$\mathbf{p}_c$, depth $d_c$, and bounds $b_c$ specifying $\Omega_c$, these
tokens define the sparse representation $\mathcal{O}(V)$.
At the finest depth, each subregion contains one voxel with zero local
coordinates, so analytic reconstruction returns its stored RGB value.
Gradients estimated from neighboring voxels are retained as encoder features.

\paragraph{Learnable residual features.}
OctVideo determines the tree topology from local approximation error.
A lightweight 3D convolutional branch extracts features
$\mathbf{F}^0=\psi_\theta(V)$ for all finest-level patches in parallel.
We then aggregate these features according to the existing tree: for each
leaf, we average and max-pool the features of the patches it covers,
concatenate the two results, and project them to a learned residual
$\mathbf{r}_c$. This adds learned detail without changing the partition.
The encoder input is
\begin{equation}
  \mathbf{x}_c^0
  =\phi_{\mathrm{tok}}(\operatorname{LN}(\mathbf{f}_c))
   +\mathbf{r}_c+\mathbf{e}_{d_c},
  \label{eq:hybrid-token}
\end{equation}
where $\mathbf{e}_{d_c}$ encodes depth. The residual projection is zero-initialized 
so that it can gradually introduce complementary information without perturbing 
the original analytic token.

\subsection{Sequence-Based Video VAE}

The VAE connects mixed-depth OctVideo cells to a compact regular latent grid.
Its encoder mixes serialized leaf features and maps them to latent cells
according to their spatio-temporal support. Its decoder starts from this grid and refines
selected regions.

\paragraph{Serialization and sequence encoding.}
Sorting the leaves by Morton keys computed from their geometric centers
defines a deterministic, locality-preserving order $\pi$ for each video.
We encode the resulting token sequence
$\mathbf{S}^0=[\mathbf{x}_{\pi(1)}^0,\ldots,\mathbf{x}_{\pi(N)}^0]$
using residual blocks:
\begin{align}
  \widetilde{\mathbf{S}}^\ell
  &=\mathbf{S}^\ell+
    \operatorname{DWConv1D}_{k,\delta_\ell}
    (\operatorname{LN}(\mathbf{S}^\ell)),\\
  \mathbf{S}^{\ell+1}
  &=\widetilde{\mathbf{S}}^\ell+
    \operatorname{MLP}(\operatorname{LN}(\widetilde{\mathbf{S}}^\ell)).
\end{align}
Small dilations extend the receptive field along the sequence. Morton order
retains coarse spatial organization, while bounds and depth preserve the
geometry needed for later alignment. For fixed width and kernel size, these
blocks have linear cost in the token count and use regular Conv1D and matrix
multiplication kernels.

\paragraph{Mapping to the latent grid.}
We use $(4,8,8)$ temporal--spatial compression, giving an
$8\times32\times32$ grid for a $32\times256\times256$ clip.
After projecting encoded features to $\mathbf{h}_c$, we copy each coarse leaf's
features to the latent cells it covers or average the features of the eight
finest-level $(2,4,4)$ siblings within a latent cell:
\begin{equation}
  \mathbf{g}_j=
  \begin{cases}
    \mathbf{h}_{o(j)}, & \text{if leaf }o(j)\text{ covers }j,\\
    \frac{1}{8}\sum_{c\in\operatorname{child}(j)}\mathbf{h}_c,
      & \text{otherwise}.
  \end{cases}
  \label{eq:copy-pool-latent}
\end{equation}
The octree bounds determine this mapping directly. We add a latent positional
embedding and a shallow RGB-to-latent residual to the resulting grid:
\begin{equation}
  \mathbf{G}^0=\mathbf{G}_{\mathrm{oct}}+\mathbf{P}_z+\mathbf{D}_r(V).
  \label{eq:dense-latent-residual}
\end{equation}
The residual branch uses a strided 3D convolution, one lightweight residual
block, and a zero-initialized output projection. A shallow grid block mixes
the combined features, and two heads then predict a diagonal Gaussian
posterior. Reparameterization yields a latent representation with 16 channels per cell. This
auxiliary path is deliberately lightweight: it adds only 30{,}849 parameters
(0.081\% of the 38.19M-parameter VAE) and 0.3 GFLOPs for a
$32\times256\times256$ clip, accounting for less than 0.2\% of the total VAE compute.

\paragraph{Selective hierarchical decoding.}
The decoder first processes the compact latent grid at $(4,8,8)$ compression with 3D convolutions.
A split head selects cells for refinement; each selected cell is projected
into eight $(2,4,4)$ child features with octant embeddings. A short Conv1D
stack mixes these children in spatial order, concentrating finer-scale
processing on the selected regions.

Each unsplit cell or refined child is mapped to local features covering its
spatio-temporal region. These features are placed at their corresponding
positions in a shared grid of size $T\times(H/2)\times(W/2)$, combining coarse
and refined regions. A lightweight 3D convolutional head then doubles the
spatial resolution and predicts the RGB video.

Training uses the input octree to supervise refinement: a latent cell is
marked for splitting when it contains finest-depth leaves. We use these
targets for teacher-forced expansion during training and predicted split
masks during validation and inference.

\paragraph{Training objective.}
The VAE training objective combines an RGB reconstruction loss, an LPIPS perceptual loss~\citep{zhang2018lpips},
Gaussian KL regularization, and split supervision:
\begin{equation}
  \mathcal{L}
  =\|\hat V-V\|_1
   +\lambda_{\mathrm{lpips}}\mathcal{L}_{\mathrm{lpips}}
   +\lambda_{\mathrm{KL}}\mathcal{L}_{\mathrm{KL}}
   +\lambda_{\mathrm{split}}\mathcal{L}_{\mathrm{split}}.
\end{equation}
Here $\mathcal{L}_{\mathrm{KL}}$ regularizes the posterior toward a standard
normal prior, and $\mathcal{L}_{\mathrm{split}}$ is binary cross-entropy on
the split logits. We set $\lambda_{\mathrm{lpips}}=0.5$ and
$\lambda_{\mathrm{split}}=0.05$. We activate the per-leaf residual $\mathbf{r}_c$ after
10\% of the total training epochs and the dense latent residual
$\mathbf{D}_r$ after 20\%. Their zero-initialized output
projections make each transition continuous; after activation, both branches
are optimized jointly with the complete VAE under $\mathcal{L}$. Full network
and training specifications are provided in
\cref{sec:appendix-architectures,sec:appendix-training-protocol}.

\subsection{Hierarchical Attention for Video Classification}

For classification, analytic tokens without the learned per-leaf residual
are grouped into patches for self-attention and feed-forward blocks.
HWT rotary position encoding provides
cell locations, and the depth embedding distinguishes cell extents. Between
stages, complete groups of eight finest-level siblings are mean-pooled to
their parent; shallower leaves pass through unchanged. The hierarchy therefore
reduces the token count as features become more semantic, without padding
the representation to a dense grid.

A single learned query then aggregates the final sparse memory
$\mathbf{M}$:
\begin{equation}
  \mathbf{g}
  =\operatorname{CrossAttn}(\mathbf{q}_{\mathrm{cls}},\mathbf{M},\mathbf{M}),
  \qquad
  \hat{\mathbf{y}}=\mathbf{W}_{\mathrm{cls}}\mathbf{g}.
\end{equation}
The query is used only at readout. We train this branch with cross-entropy.

\section{Experiments} \label{sec:result}

We first examine the compactness and signal fidelity of the adaptive
representation, then evaluate VAE reconstruction and the contributions of its
design choices. Recognition provides a complementary test of semantic
retention and learnability. The VAE uses MuonAdamW and the classifier uses
AdamW; training details are provided in \cref{sec:appendix-training-protocol}.

\begin{figure}[!t]
  \centering
  \includegraphics[width=\linewidth]{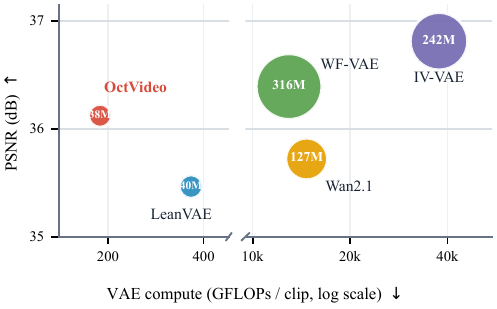}
  \caption{K400 quality--efficiency comparison. FLOPs use 32-frame clips.
  The log compute axis omits an empty range; bubble area represents parameter count.}
  \Description{A bubble plot comparing video VAEs by reconstruction PSNR,
  encoder--decoder GFLOPs, and parameter count.}
  \label{fig:vae-efficiency}
\end{figure}

\subsection{Representation Analysis}

\begin{figure*}[t]
  \centering
  \includegraphics[width=\textwidth,trim=0 223.2bp 0 0,clip]{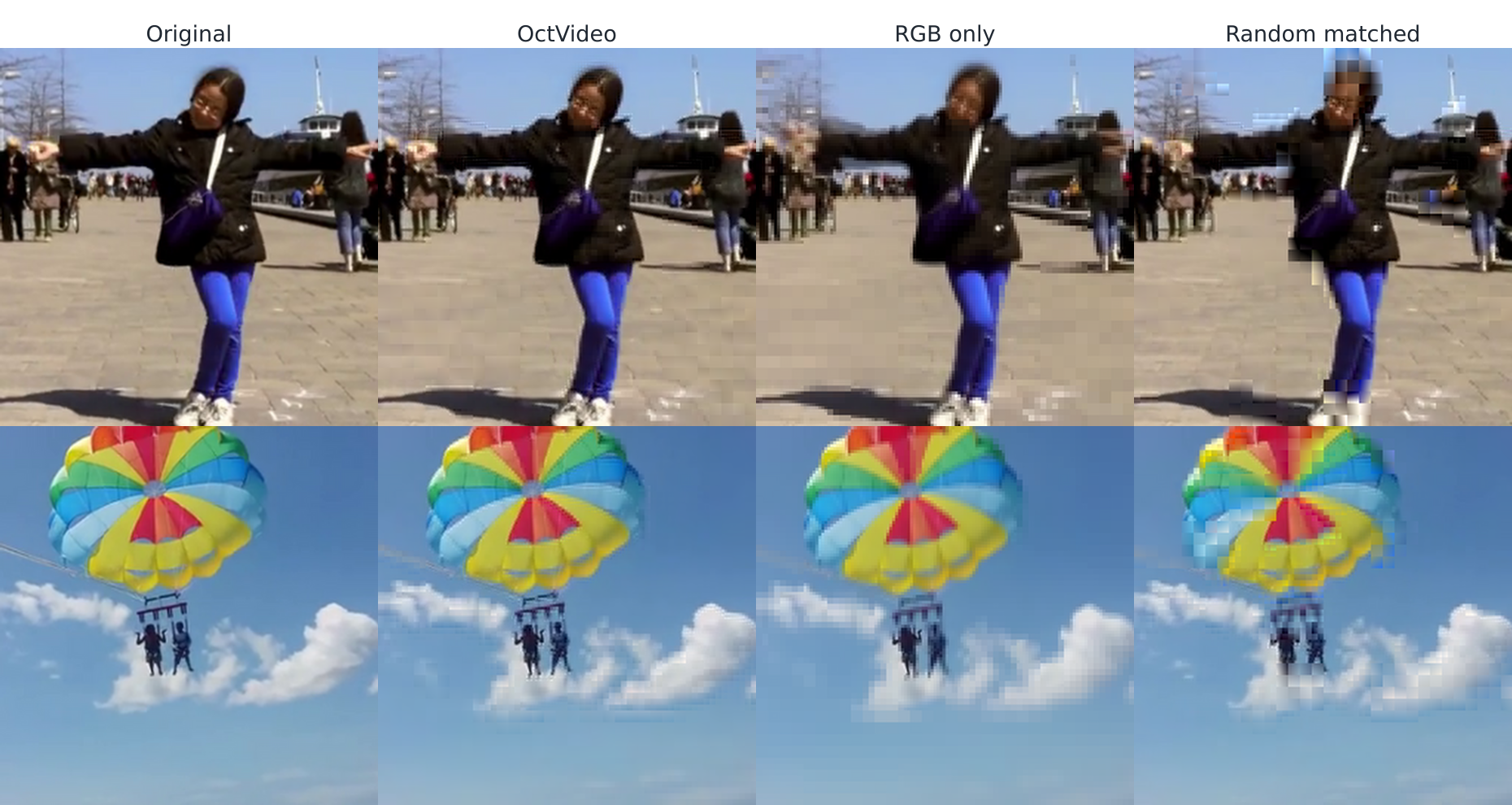}
  \caption{Direct reconstruction without a VAE: original, full OctVideo,
  RGB-only descriptors on the same tree, and a random tree matched by
  per-depth token counts. Gradients and adaptive placement preserve detail.}
  \label{fig:k400-direct-reconstruction}
\end{figure*}

\paragraph{Compactness and direct reconstruction.}
For $32\times256\times256$ clips, the finest partition has 65,536 patches.
On K400 validation, the VAE uses temporal interval 1 and averages 3,657.66
leaves ($17.9\times$ fewer); classification uses interval 2 and approximately
4,200 leaves under the same thresholds.
Each leaf retains a local
RGB approximation, allowing us to inspect the compressed signal before
training a neural decoder. Direct reconstruction achieves 34.72 dB PSNR on
K400 when PSNR is computed per clip and then averaged; details are provided in
\cref{sec:appendix-direct-reconstruction}.
Without a learned decoder, removing gradients weakens thin contours, while
randomly placing cells with matched per-depth counts introduces foreground
artifacts (\cref{fig:k400-direct-reconstruction}). Both the descriptor and
error-guided placement contribute to signal fidelity.

\paragraph{Construction-controlled signal fidelity.}
Smaller construction thresholds retain more leaves and improve direct PSNR
(\cref{fig:direct-threshold-sweep}). This tests the analytic representation
alone, unlike the fixed-VAE inference sweep in \cref{fig:threshold-sweep}.

\IfFileExists{images/direct_threshold_depth.pdf}{%
  \begin{figure}[!t]
    \centering
    \includegraphics[width=\linewidth]{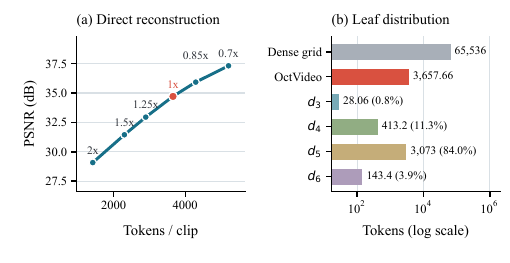}
    \caption{K400 representation analysis. (a) Analytic reconstruction
    under varying thresholds; red marks the default. (b) Token counts and
    leaf-depth shares; OctVideo averages 3,657.66 leaves.}
    \label{fig:direct-threshold-sweep}
  \end{figure}
}{%
  \noindent\begin{minipage}{\linewidth}
    \centering
    \includegraphics[width=\linewidth]{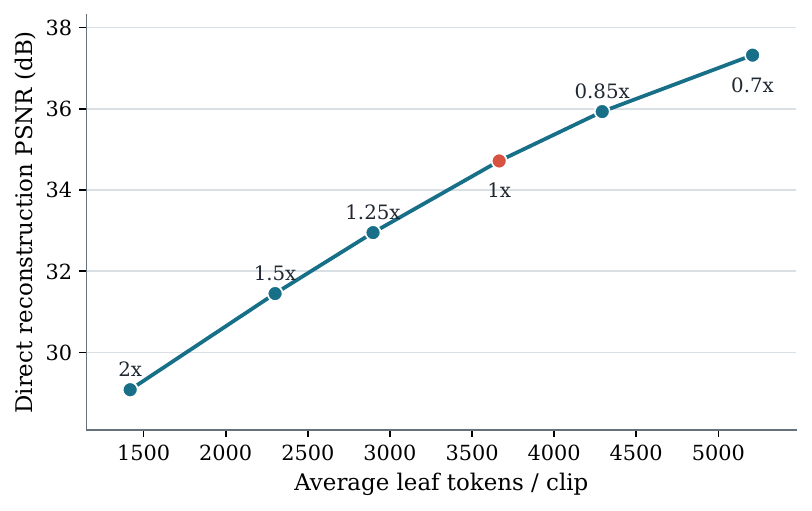}
    \captionsetup{hypcap=false}
    \captionof{figure}{Construction-threshold sweep on K400 validation without a VAE.
    The red point marks the default $(0.7,0.8,1.0)$ schedule at 34.72 dB.
    PSNR is computed per clip and then averaged.}
    \label{fig:direct-threshold-sweep}
  \end{minipage}
}

\subsection{VAE Reconstruction}

Our VAE is trained from scratch on K400 for 50 epochs with MuonAdamW,
32-frame $256^2$ clips, BF16, one clip per GPU, and eight-step accumulation.
The learning rate warms up for two epochs to $6\times10^{-4}$, then follows
cosine decay to $10^{-6}$. KL
warms up to $10^{-6}$ over 10,000 steps. EMA uses decay 0.999.
All compared VAEs use $256^2$ clips,
$4\times8\times8$ compression, and 16 latent channels. All K400 runtimes are
measured on a single RTX 4090. \Cref{tab:vae-overview} reports encoding/decoding
only; highly parallelized OctVideo construction takes under 2 ms, and its
end-to-end construction, encoding, and decoding take approximately 22 ms per video.

\begin{table*}[t]
  \tablestyle{1.5pt}{1.02}
  \caption{K400 video VAE comparison at $256^2$ with $4\times8\times8$
  compression and 16 latent channels. All models use 32-frame clips.
  PSNR is computed per clip and then averaged.
  Memory: peak GiB;
  runtime: encoder/decoder seconds.}
  \label{tab:vae-overview}
  \begin{tabular*}{\textwidth}{@{\extracolsep{\fill}}lrrrrrrrrrr@{}}
    \hline
    Method & $T$ & Params & GFLOPs $\downarrow$ & Mem. $\downarrow$ & MSE $\downarrow$ &
    MAE $\downarrow$ & PSNR $\uparrow$ & SSIM $\uparrow$ & LPIPS $\downarrow$ & Enc./Dec. \\
    \hline
    Wan2.1 & 32 & 127M & 14{,}685.8 & 1.669 & $4.65{\times}10^{-4}$ & 0.0108 & 35.72 & 0.958 & \best{0.0182} & 0.2005/0.3423 \\
    CogVideoX & 32 & 206M & 34{,}360.1 & 4.924 & $1.12{\times}10^{-3}$ & 0.0148 & 32.30 & 0.939 & 0.0380 & 0.4403/0.8770 \\
    IV-VAE & 32 & 242M & 37{,}662.3 & 2.082 & $\bestmath{3.74{\times}10^{-4}}$ & \best{0.0096} & \best{36.81} & \best{0.966} & 0.0188 & 0.9076/0.8304 \\
    WF-VAE & 32 & 316M & 12{,}952.8 & 2.693 & $4.28{\times}10^{-4}$ & 0.0097 & 36.39 & 0.964 & 0.0226 & 0.0700/0.2361 \\
    LeanVAE & 32 & 40M & 365.1 & \best{0.280} & $4.97{\times}10^{-4}$ & 0.0109 & 35.46 & 0.956 & 0.0200 & 0.0116/0.0207 \\
    \textbf{OctVideo} & 32 & \best{38M} & \best{188.9} & 0.803 & $4.83{\times}10^{-4}$ & 0.0106 & 36.12 & 0.960 & 0.0220 & \best{0.0105/0.0091} \\
    \hline
  \end{tabular*}
\end{table*}

\paragraph{Reconstruction quality and efficiency.}
Among the compared VAEs in \cref{tab:vae-overview,fig:vae-efficiency}, OctVideo uses the fewest
parameters and requires the fewest FLOPs, with the shortest encoding and decoding
times, while achieving high-quality reconstruction at 36.12 dB PSNR,
0.960 SSIM, and 0.0220 LPIPS.

\begin{figure}[t]
  \centering
  \includegraphics[width=\linewidth]{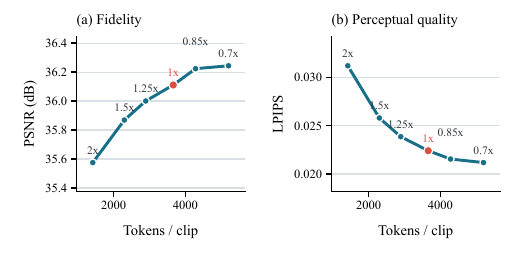}
  \caption{Test-time token scaling on the full K400 validation set with a
  fixed VAE checkpoint. Labels indicate threshold multipliers; red marks
  the training setting ($1\times$).}
  \Description{Reconstruction PSNR and LPIPS versus mean input tokens as
  construction thresholds vary with fixed VAE weights.}
  \label{fig:threshold-sweep}
\end{figure}

\paragraph{Test-time token scaling.}
With VAE weights fixed, lowering the construction thresholds retains more
input tokens and improves PSNR and LPIPS, with diminishing gains over the
evaluated range (\cref{fig:threshold-sweep}). This provides a quality--token
trade-off without retraining the VAE.

\subsubsection{Zero-Shot DAVIS Reconstruction}

We directly evaluate the K400-trained OctVideo VAE on all 50 DAVIS 2016
sequences at their original resolution~\citep{perazzi2016davis}, without
fine-tuning or model selection on DAVIS. All methods use 32-frame clips
under native or tiled high-resolution inference. The comparison spans
industrial and academic VAEs of different scales. PSNR is computed per clip
and then averaged equally across clips; evaluation details are in
\cref{sec:appendix-highres-protocol}.

\begin{table}[t]
  \tablestyle{0.8pt}{1.08}
  \caption{Zero-shot reconstruction on all 50 DAVIS 2016 sequences at native
  resolution. $T$: input frames; time: VAE seconds per clip; compute:
  mean TFLOPs ($1\ \mathrm{MAC}=1\ \mathrm{FLOP}$).}
  \label{tab:davis-zero-shot}
  \begin{tabular}{@{}lrrrrrrr@{}}
    \hline
    Method & $T$ & PSNR $\uparrow$ & SSIM $\uparrow$ &
    LPIPS $\downarrow$ & rFVD $\downarrow$ & Time $\downarrow$ & TFLOPs$\downarrow$ \\
    \hline
    Wan2.1 & 32 & \best{33.47} & 0.904 & \best{0.071} & 23.0 & $>20$ & 466.89 \\
    HunyuanVideo & 32 & 32.63 & \best{0.909} & 0.080 & 27.1 & $>20$ & 1484.95 \\
    LeanVAE & 32 & 33.07 & 0.899 & 0.076 & \best{18.8} & 0.820 & 14.25 \\
    \textbf{OctVideo} & 32 & 32.98 & 0.894 & 0.093 & 20.3 & \best{0.540} & \best{8.47} \\
    \hline
  \end{tabular}
\end{table}

\begin{figure*}[t]
  \centering
  \includegraphics[width=\textwidth]{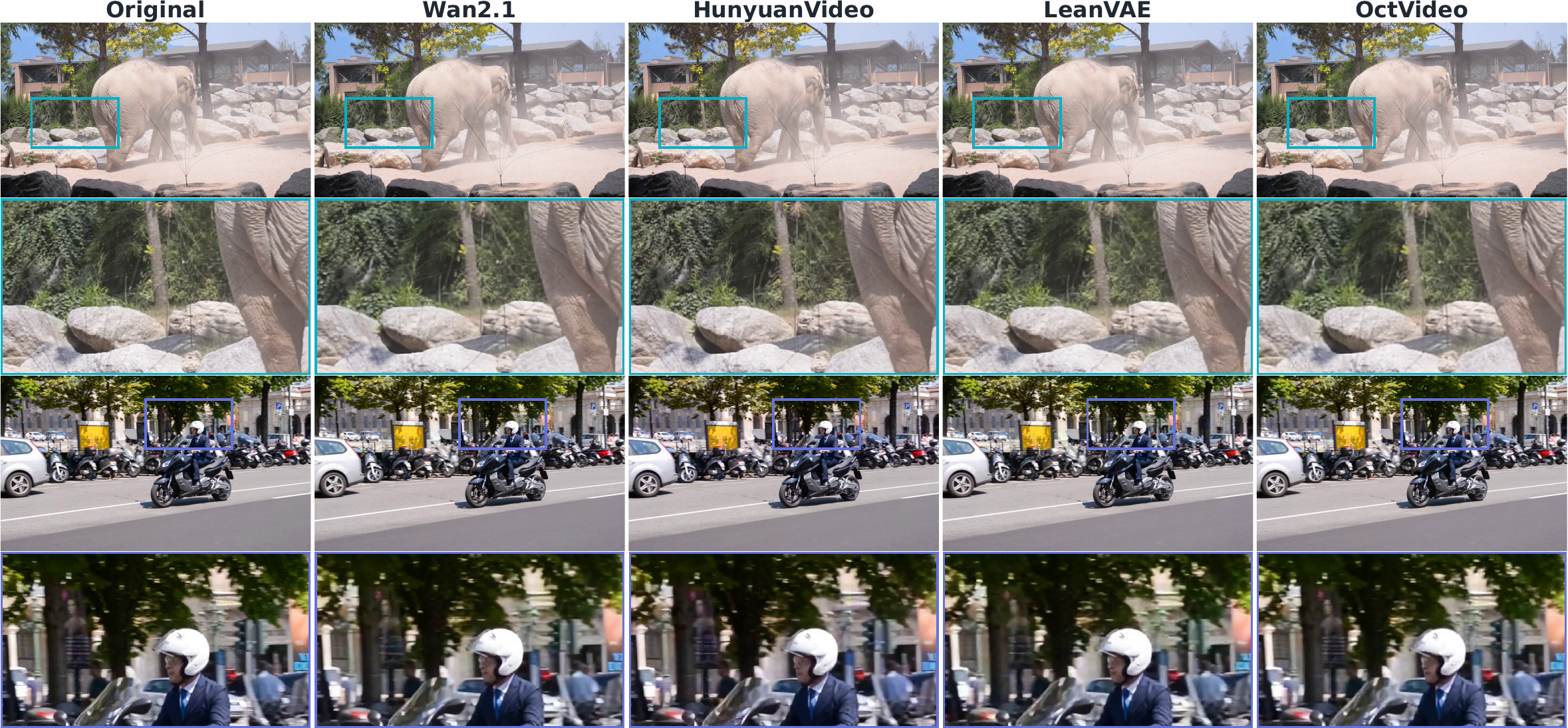}
  \caption{Qualitative reconstruction comparison on selected DAVIS 2016
  examples.}
  \Description{Aligned DAVIS video frames and enlarged local regions comparing
  the original input with Wan2.1, HunyuanVideo, LeanVAE, and OctVideo
  reconstructions.}
  \label{fig:davis-qualitative}
\end{figure*}

OctVideo achieves substantially lower VAE runtime while preserving global
structure and local details, yielding visually competitive reconstructions in
\cref{fig:davis-qualitative}.

\subsection{VAE Ablation Studies}

All variants are trained from scratch on split~1 of the UCF101
action-recognition dataset~\citep{soomro2012ucf101} with the same data,
latent size, optimization, and model-selection protocol. These controls
examine the contributions of the local descriptor, adaptive cell placement,
and neural processing to VAE quality and cost.

\begin{table}[t]
  \tablestyle{2.5pt}{1.06}
  \caption{Controlled VAE ablations on all 3,783 UCF101 split-1 test
  clips, with the same latent size and training recipe. Dense Conv3D inputs
  use 8,192 tokens; sparse inputs use 3,225.
  $1\ \mathrm{MAC}=1\ \mathrm{FLOP}$.}
  \label{tab:ucf101-vae-core-ablation}
  \begin{tabular}{@{}lrrrr@{}}
    \hline 
    Variant & GFLOPs & PSNR $\uparrow$ & SSIM $\uparrow$ & LPIPS $\downarrow$ \\
    \hline
    Dense Conv3D (small) & 239.7 & 32.30 & 0.928 & 0.0605 \\
    Dense Conv3D (large) & 1,780.5 & 34.12 & 0.947 & 0.0399 \\
    RGB-only tokens & 181.2 & 32.70 & 0.932 & 0.0537 \\
    Random matched tree & 181.3 & 33.50 & 0.941 & 0.0422 \\
    Dense decoder & 835.9 & 33.90 & 0.940 & 0.0403 \\
    w/o latent residual & 181.0 & 32.20 & 0.913 & 0.0535 \\
    Latent residual only & \textbf{126.3} & 31.87 & 0.897 & 0.0685 \\
    \textbf{OctVideo} & 181.3 & \textbf{34.18} & \textbf{0.948} & \textbf{0.0382} \\
    \hline
  \end{tabular}
\end{table}

\begin{table*}[!t]
  \tablestyle{1.5pt}{1.0}
  \caption{Action recognition on Kinetics-400. Token counts indicate the per-view
  sequence length after initial embedding; OctVideo reports the mean number of adaptive
  leaves.}
  \label{tab:k400-classification}
  \begin{tabular}{lccccccc}
    \hline
    Method & Pre-train & Params & Input & Views & Tokens/view & Top-1 & Top-5 \\
    \hline
    Rev-MViT-B~\citep{mangalam2022reversible} & none & 34.9M & $16{\times}224^2$ & $5{\times}1$ & 25{,}088 & 78.5 & 93.4 \\
    MViTv1-B~\citep{fan2021multiscale} & none & 36.6M & $32{\times}224^2$ & $5{\times}1$ & 50{,}176 & 80.2 & 94.4 \\
    MViTv2-S~\citep{li2022mvitv2} & none & 34.5M & $16{\times}224^2$ & $5{\times}1$ & 25{,}088 & 81.0 & 94.6 \\
    \textbf{OctVideo} & none & 33.0M & $32{\times}256^2$ & $4{\times}3$ & $\sim$4{,}200 & 80.1 & 94.4 \\
    \hline
    Video Swin-T~\citep{videoswinT} & IN-1K & 28.2M & $32{\times}224^2$ & $4{\times}3$ & 50{,}176 & 78.8 & 93.6 \\
    Hiera-B~\citep{ryali2023hiera} & K400 MAE & 51.0M & $16{\times}224^2$ & $5{\times}3$ & 25{,}088 & 84.0 & -- \\
    VideoMAE-S~\citep{tong2022videomae} & K400 SSL & 22.0M & $16{\times}224^2$ & $5{\times}3$ & 1{,}568 & 79.0 & 93.8 \\
    \hline
  \end{tabular}
\end{table*}

\paragraph{Adaptive structure improves reconstruction.}
The small Conv3D baseline is the closest dense alternative in computational cost, yet
it trails OctVideo by 1.88 dB and has substantially worse LPIPS. Increasing its
capacity recovers nearly the same reconstruction quality (34.12 versus 34.18
dB), but raises computation from 239.7 to 1{,}780.5 GFLOPs. Among the dense
baselines evaluated, increasing capacity approaches OctVideo's reconstruction
quality at substantially higher computational cost.

\paragraph{The analytic descriptor matters.}
Removing the explicit spatio-temporal gradients while preserving the adaptive
tree, token width, and network lowers PSNR by 1.48 dB and degrades both SSIM
and LPIPS. The gradient channels are therefore useful signal priors rather
than redundant inputs that the learned residual can easily replace.

\paragraph{The latent residual complements the sparse pathway.}
Removing the dense latent residual reduces PSNR by 1.98 dB, while using this
branch alone reaches 31.87 dB, 2.31 dB below the complete model. Its much
higher LPIPS (0.0685 versus 0.0382) further shows that this dense branch
alone cannot match the full model's reconstruction quality.
The two pathways therefore provide complementary reconstruction information.

\paragraph{Error-guided allocation improves reconstruction.}
The random-tree control retains the same RGB+gradient tokens and per-depth
counts, but places refinements independently of approximation error;
its 0.68 dB drop in PSNR shows that sparsity alone is insufficient.
Replacing our selective decoder with a dense 3D convolutional decoder requires
$4.6\times$ the total VAE FLOPs and yields 0.28 dB lower PSNR.
These controls support error-guided input allocation and the efficiency of
the proposed decoder.

\subsection{Latent Video Generation}
\label{sec:generation}

\begin{figure}[!ht]
  \centering
  \includegraphics[width=\linewidth]{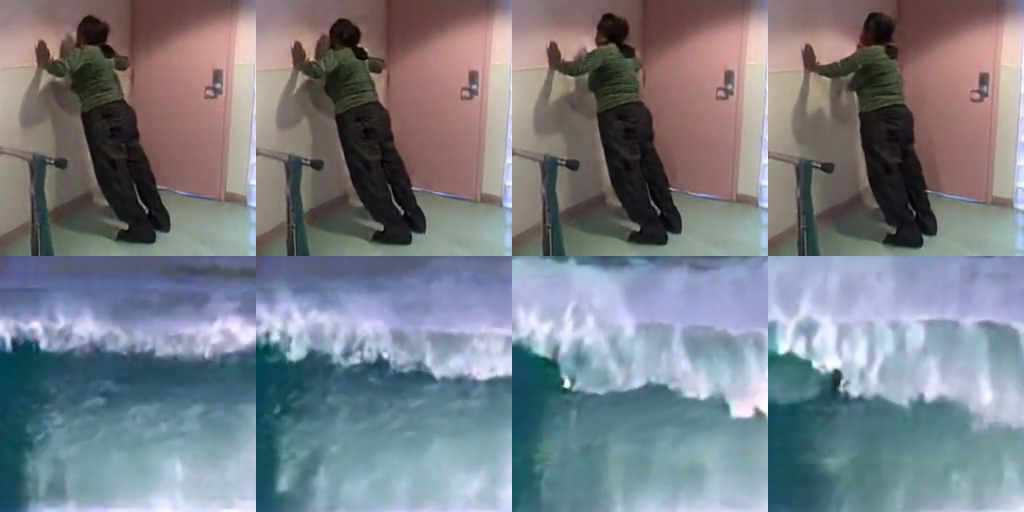}
  \caption{Class-conditional samples: WallPushups (top) and Surfing
  (bottom). Each row shows a selected excerpt at 0.25-second intervals.}
  \label{fig:latent-generation}
\end{figure}

The regular latent grid directly supports standard generators without
additional position modeling. Keeping the K400-trained VAE fixed, we train
a class-conditional Latte-style Transformer~\citep{ma2025latte,peebles2023dit}
with approximately 674M parameters from scratch on UCF101
split~1~\citep{soomro2012ucf101}, using flow matching~\citep{lipman2023flowmatching}
on its latents. After a compute-limited budget of 150,000 steps, evaluation
on 5,050 generated clips gives FVD 264 and KVD 28.6~\citep{unterthiner2018fvd}
(\cref{tab:latent-generation}). Samples are shown in \cref{fig:latent-generation};
protocol details are in \cref{sec:appendix-generation}.

\begin{table}[!ht]
  \tablestyle{3pt}{1.06}
  \caption{Class-conditional UCF101 generation: 29 frames at $256^2$
  and 12 fps. Parameters refer to the generator.}
  \label{tab:latent-generation}
  \begin{tabular}{@{}lrrrr@{}}
    \hline
    Generator & Params & Steps & FVD $\downarrow$ & KVD $\downarrow$ \\
    \hline
    Latte + FM & $\sim$674M & 150K & 264 & 28.6 \\
    \hline
  \end{tabular}
\end{table}

\subsection{Video Understanding}

As a complementary evaluation beyond reconstruction, we test whether the
adaptive representation retains semantic information and supports learned
recognition.

\paragraph{Semantic retention under direct reconstruction.}
A frozen Video Swin-B evaluator, pretrained on ImageNet-22K and fine-tuned
on K400, evaluates original videos and their analytic OctVideo reconstructions
(\cref{tab:direct-recon-videoswin}). It is distinct from the ImageNet-1K-pretrained
Swin-T baseline in \cref{tab:k400-classification}. Top-1 accuracy decreases
from 82.27\% to 79.51\%. Because neither a VAE nor adaptation to the
reconstructed inputs is involved, this measures the effect of the
representation on the frozen evaluator.

\begin{table}[!htbp]
  \tablestyle{2pt}{1.08}
  \caption{Semantic preservation under direct OctVideo reconstruction. A
  frozen Video Swin-B (IN-22K pretraining, K400 fine-tuning) evaluates both inputs.}
  \label{tab:direct-recon-videoswin}
  \begin{tabular}{lcccc}
    \hline
    Input & Top-1 & Top-5 & $\Delta$Top-1 & $\Delta$Top-5 \\
    \hline
    Original video & 82.27 & 96.80 & -- & -- \\
    Direct reconstruction & 79.51 & 94.40 & $-2.76$ & $-2.40$ \\
    \hline
  \end{tabular}
\end{table}

\paragraph{Recognition learned on OctVideo.}
Parameter-free sibling pooling reduces the approximately 4{,}200 input tokens
to 3{,}700 and 800 across classifier stages (supplementary
Table~\ref{tab:stage-tokens}).
We train the classifier from scratch on Kinetics-400 without external
pre-training. It receives $32$ frames at temporal interval 2 and $256\times256$ resolution and is
evaluated with four temporal clips and three spatial crops; token counts below
are for one view.

With 33.0M parameters and approximately 4{,}200 input tokens, OctVideo reaches
80.1\% Top-1 and 94.4\% Top-5 accuracy. Transferring the K400-trained classifier to UCF101
split~1 yields 96.43\% and 99.52\%, respectively; details are in
supplementary Table~\ref{tab:ucf101-baselines}.

\section{Conclusion} \label{sec:conclusion}

We presented OctVideo, an adaptive spatio-temporal representation in which
local signal approximation determines cell refinement. RGB values and
gradients describe the video within each cell, preserving coarse descriptions
of smooth regions alongside finer ones for detailed regions. Learned residual features
supplement these descriptors for neural processing. Sequence-based Conv1D
encoding, alignment to a regular latent grid, and selective refinement make
the representation practical for video reconstruction. Experiments on
Kinetics-400 and zero-shot transfer to DAVIS demonstrate competitive pixel-level
fidelity at low computational cost. Controlled UCF101 ablations support the
design choices: naive dense scaling is
costly, explicit gradients improve the token signal, and error-guided input
and decoder topologies outperform unselective alternatives. Complementary recognition experiments show that the same
adaptive cells retain semantic information and support learned classification
through patch-wise attention and sibling pooling. A compute-limited UCF101
experiment further demonstrates that OctVideo latents support video generation.

These results show that compact signal descriptions can also structure neural
computation. Future work includes improving fine spatio-temporal detail and
exploiting fast decoding in repeated sampling and evaluation. Integrating
OctVideo into vision-language models could further support efficient video processing.

\FloatBarrier
{\small
\bibliographystyle{ieeenat_fullname}
\bibliography{main}
}
\clearpage
\appendix
\onecolumn
\begin{center}
{\Large\bfseries\thetitle\par}
\medskip
{\large Supplementary Material\par}
\end{center}
\pdfbookmark[0]{Supplementary Material}{supplementary}
\suppressfloats[t]
\section{Appendix} \label{sec:appendix}

\subsection{Video Understanding Details}
\label{sec:appendix-token-distribution}

\Cref{tab:stage-tokens} reports how the adaptive input is reduced across the
recognition hierarchy. Pooling is parameter-free: complete groups of eight
children are replaced by their parent, while shallower leaves pass through.

\begin{table}[!htbp]
  \tablestyle{4pt}{1.08}
  \caption{K400 recognition token hierarchy for one
  $32\times256\times256$ view sampled at temporal interval 2. Rows after the
  sparse input are classifier states produced by sibling pooling.}
  \label{tab:stage-tokens}
  \begin{tabular}{lcccc}
    \hline
    Level & Active depths & Operation & Consumer & Avg. tokens \\
    \hline
    Sparse input & $d_3$--$d_6$ & adaptive leaves & classifier & $\sim$4{,}200 \\
    Level-5 state & $d_3$--$d_5$ & $d_6\!\rightarrow\!d_5$ pool & classifier & $\sim$3{,}700 \\
    Level-4 state & $d_3$--$d_4$ & $d_5\!\rightarrow\!d_4$ pool & classifier & $\sim$800 \\
    \hline
  \end{tabular}
\end{table}
\FloatBarrier

\paragraph{UCF101 transfer.}
Starting from the K400-supervised OctVideo checkpoint, we replace the
classification head and fine-tune on the official UCF101
split~1~\citep{soomro2012ucf101}. No UCF101 video is used in pre-training.
Fine-tuning uses AdamW for 50 epochs with a base learning rate of
$3\times10^{-5}$ and weight decay of 0.05.

\begin{table}[!htbp]
  \tablestyle{3pt}{1.05}
  \caption{UCF101 transfer and representative RGB-based methods. Published
  methods retain their original protocols; OctVideo uses four temporal clips
  and three spatial crops on official split~1.}
  \label{tab:ucf101-baselines}
  \begin{tabular}{lccc}
    \hline
    Method & Pretraining & Top-1 & Top-5 \\
    \hline
    C3D (3 nets)~\citep{tran2015learning} & I380K+Sports-1M & 85.2 & -- \\
    I3D-RGB~\citep{Carreira_2017_CVPR} & ImageNet+K400 & 95.6 & -- \\
    R(2+1)D-RGB~\citep{tran2018closer} & K400 & 96.8 & -- \\
    TSM~\citep{lin2019tsm} & K400 & 95.9 & 99.7 \\
    VideoMAE V2-g~\citep{wang2023videomaev2} & UnlabeledHybrid+K710 & \textbf{99.6} & \textbf{100.0} \\
    \textbf{OctVideo (ours)} & K400 & 96.43 & 99.52 \\
    \hline
  \end{tabular}
\end{table}
\FloatBarrier

\subsection{Training and Evaluation Protocols}
\label{sec:appendix-training-protocol}
\label{sec:appendix-highres-protocol}

\paragraph{Optimization.}
The K400 classifier uses AdamW with $\beta=(0.9,0.999)$ and weight decay
0.05. It is trained for 200 epochs at a base learning rate of $10^{-4}$,
with 30 epochs of linear warm-up from $0.01\times$ the base learning rate, followed by cosine decay
to $10^{-6}$. The per-GPU batch size is two with two-step gradient accumulation.
Gradients are clipped to norm 1.0, stochastic depth increases linearly from
0 to 0.2, and an exponential moving average with rate 0.999 is maintained.
The VAE uses MuonAdamW: Muon updates large matrix parameters and AdamW
updates the remaining parameters. It is trained for 50 epochs at
$6\times10^{-4}$, with a two-epoch warm-up from $0.1\times$ the base learning rate and cosine decay
to $10^{-6}$, using one clip per GPU and eight-step gradient accumulation.
Muon uses momentum 0.95, five Newton--Schulz steps, Nesterov momentum, and
weight decay 0.05; AdamW uses $\beta=(0.9,0.999)$ and $\epsilon=10^{-8}$.
Both models use BF16 automatic mixed precision. The VAE maintains an
exponential moving average with rate 0.999 and is trained without an
adversarial loss.

\paragraph{Video generation.}
The experiment in \cref{sec:generation} uses the same K400-trained OctVideo
VAE without updating its parameters. The full generation protocol is
provided in \cref{sec:appendix-generation}.

\paragraph{Classification augmentation.}
We follow the MViT augmentation recipe~\citep{fan2021multiscale,li2022mvitv2},
including random spatial crops and horizontal flips, RandAugment, Mixup,
CutMix, and random erasing. Specifically, we use two augmented views per
training video, four RandAugment operations of magnitude 7, color jitter with
probability 0.8 and magnitude 0.4, random erasing with probability 0.25,
Mixup $\alpha=0.8$, CutMix $\alpha=1.0$, and label smoothing 0.1. Mixup or
CutMix is applied with probability 1.0, with equal probability of selecting
either operation.

\paragraph{High-resolution reconstruction.}
For the DAVIS comparison in \cref{tab:davis-zero-shot}, we extract one
center-aligned 32-frame clip from each of the 50 sequences for every model.
OctVideo uses the VAE trained on K400 directly, without DAVIS fine-tuning;
baseline VAE parameters are also kept unchanged during evaluation.
Wan2.1 uses native full-frame inference (32 frames, internally
padded to 33, with four-GPU layer-wise parallelism). HunyuanVideo, LeanVAE,
and OctVideo use non-overlapping $256\times256$ tiles.
Inputs are replicate-padded when needed, and outputs are cropped after
full-frame reconstruction or tile stitching.

External-VAE FLOPs count encoder and decoder convolutions and matrix products
at their executed shapes, including internal padding to 33 frames, with
$1$ MAC counted as $1$ FLOP. Normalization, activations, interpolation, and
metric networks are excluded. The 46 sequences at $1920\times1080$ and four
at $1600\times900$ average 39.04 tiles per clip, including edge padding.
Wan is counted at each padded full-frame resolution instead.

Metrics are computed on all valid aligned frames after restoring the original
resolution and are averaged equally across sequences. For both K400 and DAVIS,
PSNR is computed from the RGB $[0,1]$ MSE over each complete clip and then
averaged equally across clips, rather than averaging frame-wise PSNR or
converting the dataset-mean MSE to PSNR.
SSIM~\citep{wang2004ssim} also uses RGB values in $[0,1]$,
LPIPS~\citep{zhang2018lpips} uses the AlexNet backbone for every method, and rFVD uses one
Kinetics-400 I3D feature per aligned clip.

\subsection{Network Architectures}
\label{sec:appendix-architectures}

\begin{figure}[t]
  \centering
  \includegraphics[width=\textwidth]{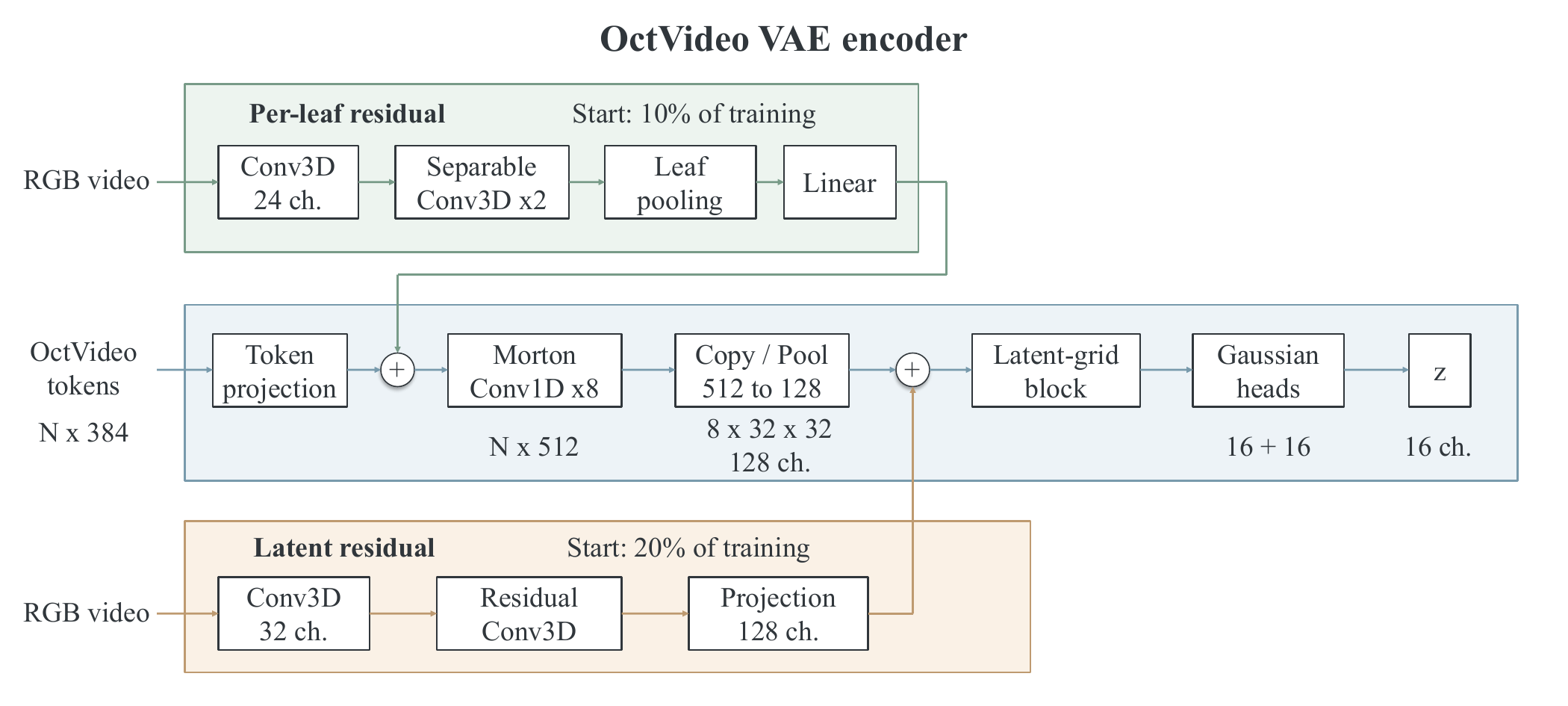}
  \caption{OctVideo VAE encoder. The per-leaf branch pools finest-level
  features according to the OctVideo partition and adds them before sequence
  encoding. The latent residual is added after copy/pool mapping and before
  the latent-grid block. The branches are enabled after 10\% and 20\% of the
  total training epochs, respectively. Gaussian heads predict the mean and
  log variance; $z$ is sampled during training and uses the mean at evaluation.
  Normalization, activations, and positional embeddings are omitted for clarity.}
  \label{fig:appendix-vae-encoder}
\end{figure}

\paragraph{VAE architecture.}
\Cref{fig:appendix-vae-encoder} summarizes the encoder and the two residual
branches.
\Cref{tab:vae-architecture} gives the complete architecture used in our main
experiments. Every leaf stores a $2\times4\times4$ grid of 12-dimensional
RGB-and-gradient descriptors, flattened into one 384-dimensional token.
A depth-$d_c$ leaf covers $(2s,4s,4s)$ video samples, with
$s=2^{D-d_c}$; only at the finest depth $D$ is its coverage $2\times4\times4$.

The per-leaf residual branch first applies a 3D convolution from 3 to 24
channels with kernel and stride $(2,4,4)$, followed by SiLU. Two separable
3D layers each use a $3^3$ depthwise convolution and a $1^3$ pointwise
convolution, with SiLU after both convolutions. We recursively construct a
feature pyramid by $2^3$ pooling and concatenate the mean and maximum over
each leaf support. A zero-initialized linear layer maps the resulting 48
features to 512 channels. Its learnable scalar is initialized to one, and the
branch is activated after 10\% of the total training epochs.

The analytic token is processed by LayerNorm, a $384\!\rightarrow\!512$
linear layer, and GELU; the per-leaf residual and a learned depth embedding
are then added. Eight pre-normalized Morton Conv1D blocks use width 512,
depthwise kernel 7, dilations $(1,2,4,1,2,4,1,1)$, and an
$512\!\rightarrow\!2048\!\rightarrow\!512$ GELU MLP. Copy/pool resampling
first projects features from 512 to 128 channels. Coarse leaves are copied to
every covered latent location, whereas each complete group of eight
finest-depth siblings is averaged into one location, producing an
$8\times32\times32\times128$ grid.

The dense latent residual applies a 3D convolution from 3 to 32 channels with
kernel and stride $(4,8,8)$ and SiLU. It then uses one residual block composed
of GroupNorm, a $3^3$ depthwise convolution, SiLU, and a $1^3$ pointwise
convolution. A zero-initialized $1^3$ convolution projects to 128 channels;
its scale is initialized to one and the branch is activated after 20\% of the
total training epochs. Its
output is added to the copy/pool grid before the latent-grid block. The latter
uses GroupNorm, a $3^3$ depthwise convolution, and a
$128\!\rightarrow\!128\!\rightarrow\!128$ pointwise GELU MLP, all with
residual connections. After LayerNorm, separate linear heads predict 16
means and 16 log variances per grid location.

\begin{table}[t]
  \centering
  \caption{Layer-by-layer specification of the OctVideo VAE. Shapes omit the
  batch dimension. DW denotes a depthwise convolution.}
  \label{tab:vae-architecture}
  \footnotesize
  \setlength{\tabcolsep}{3pt}
  \renewcommand{\arraystretch}{1.08}
  \begin{tabular}{p{0.14\textwidth}p{0.20\textwidth}p{0.43\textwidth}p{0.15\textwidth}}
    \hline
    Component & Input $\rightarrow$ output & Layers and hyperparameters & Activation / normalization \\
    \hline
    Analytic token projection & $N\times384\rightarrow N\times512$ & Linear $384\rightarrow512$ & LayerNorm, GELU \\
    Per-leaf residual & $3\times32\times256\times256\rightarrow N\times512$ & Conv3D $(2,4,4)$, 24 ch.; $2\times$[DWConv3D $3^3$, Conv3D $1^3$]; mean+max pool; Linear $48\rightarrow512$ & SiLU; zero-init. output; start at 10\% \\
    Morton sequence encoder & $N\times512\rightarrow N\times512$ & 8 blocks; DWConv1D $k=7$; dilations $1,2,4,1,2,4,1,1$; MLP ratio 4 & Pre-LayerNorm, GELU \\
    Copy/pool mapping & $N\times512\rightarrow8\times32\times32\times128$ & Linear $512\rightarrow128$; copy coarse leaves / mean-pool 8 children & learned 3D position embedding \\
    Dense latent residual & $3\times32\times256\times256\rightarrow8\times32\times32\times128$ & Conv3D $(4,8,8)$, 32 ch.; 1 residual [DWConv3D $3^3$, Conv3D $1^3$]; Conv3D $1^3$ to 128 & GroupNorm, SiLU; zero-init. output; start at 20\% \\
    Latent-grid block & $8\times32\times32\times128$ & 1 block; DWConv3D $3^3$; pointwise MLP ratio 1 & GroupNorm, GELU \\
    Posterior heads & $8192\times128\rightarrow8192\times(16+16)$ & separate Linear $128\rightarrow16$ for $\boldsymbol\mu$ and $\log\boldsymbol\sigma^2$ & LayerNorm \\
    Latent decoder stem & $8192\times16\rightarrow8\times32\times32\times384$ & Linear $16\rightarrow384$; learned 3D position embedding & -- \\
    Coarse decoder & $8\times32\times32\times384$ & 3 residual blocks; Conv3D $3^3$; MLP $384\rightarrow768\rightarrow384$ & LayerNorm, SiLU/GELU \\
    Split and expansion & coarse grid $\rightarrow$ selected $16\times64\times64$ cells & LN--Linear $384\rightarrow1$; Linear $384\rightarrow8\times384$; 8 octant embeddings & sigmoid threshold 0.5 \\
    Selective refinement & selected cells, 384 ch. & 2 Morton DWConv1D blocks, $k=7$, dilations $(1,2)$; MLP ratio 2 & Pre-LayerNorm, GELU \\
    Canvas fusion & coarse/refined cells $\rightarrow24\times32\times128\times128$ & Linear $384\rightarrow24\cdot4^3$ (coarse) or $384\rightarrow24\cdot2^3$ (refined) & coordinate placement \\
    RGB head & $24\times32\times128\times128\rightarrow3\times32\times256\times256$ & DWConv3D $3^3$; Conv3D $1^3$, $24\rightarrow32$; nearest spatial $2\times$; DWConv3D $3^3$; Conv3D $3^3$, $32\rightarrow3$ & SiLU \\
    \hline
  \end{tabular}
\end{table}

\paragraph{Dense decoder ablation.}
This variant retains the encoder architecture and latent specification but
replaces the selective decoder with a dense 3D convolutional decoder.
It uses a 384-channel stem and three upsampling stages with 192, 96, and 48
channels, with two residual blocks at the stem and each stage.
The decoder uses full $3\times3\times3$ convolutions, RMSNorm, and
nearest-neighbor upsampling, without a split head or split supervision.

\paragraph{VAE objective and optimization.}
The reconstruction term is RGB $\ell_1$. LPIPS uses the AlexNet backbone and
16 randomly sampled frames per clip, with weight 0.5. The KL weight is
linearly increased from zero to $10^{-6}$ over the first 10{,}000 optimizer steps. Split logits
use binary cross-entropy with weight 0.05; the positive-class weight is
estimated per batch and capped at 64. During training, ground-truth split
masks determine expansion; inference uses a sigmoid threshold of 0.5.
Optimization begins with the octree pathway alone. The
per-leaf and latent residual branches are activated after 10\% and 20\% of
the total training epochs, respectively, and are then trained jointly with
all other VAE parameters.

\paragraph{Classification network.}
The classifier receives the same 384-dimensional analytic OctVideo tokens;
the dense residual branch is disabled for this experiment. A linear layer
maps tokens to width 512, and a two-layer
$1\!\rightarrow\!512\!\rightarrow\!512$ SiLU MLP embeds normalized octree
depth. The ten pre-LayerNorm transformer blocks are divided into stages of
$(4,4,2)$ blocks. Each block uses 8 attention heads (64 channels per head),
3D rotary position encoding, QKV bias, attention over contiguous groups of
256 hierarchy-ordered tokens, and a
$512\!\rightarrow\!2048\!\rightarrow\!512$ GELU MLP. Attention and projection
dropout are zero; stochastic depth grows linearly from 0 to 0.2 across the ten
blocks. After stages one and two, complete groups of eight depth-6 and depth-5
siblings, respectively, are averaged into their parents; unmatched shallower
leaves pass through unchanged. A final FP32 LayerNorm is followed by a learned
single-query attention readout with independent 512-dimensional query, key,
and value projections, and a $512\!\rightarrow\!400$ classifier.

\begin{table}[t]
  \tablestyle{3pt}{1.04}
  \caption{OctVideo classification architecture on K400.}
  \label{tab:classifier-architecture}
  \begin{tabular}{lccc}
    \hline
    Stage & Blocks & Token operation & Feature configuration \\
    \hline
    Input & -- & Linear $384\rightarrow512$ + depth embedding & width 512 \\
    Stage 1 & 4 & patch attention, then $d_6\rightarrow d_5$ pool & 8 heads, chunk 256, MLP $4\times$ \\
    Stage 2 & 4 & patch attention, then $d_5\rightarrow d_4$ pool & 8 heads, chunk 256, MLP $4\times$ \\
    Stage 3 & 2 & patch attention & 8 heads, chunk 256, MLP $4\times$ \\
    Readout & -- & single-query attention + Linear $512\rightarrow400$ & final FP32 LayerNorm \\
    \hline
  \end{tabular}
\end{table}

Recognition and reconstruction use the same thresholds and tree-construction
rule, but not the same temporal sampling protocol. Recognition samples 32
frames at interval 2 and averages approximately 4,200 input leaves, whereas
the VAE uses 32 consecutive frames (interval 1) and averages 3,657.66 leaves on
K400 validation. The larger recognition count therefore reflects
the greater temporal span and variation of its input, rather than data
augmentation or a different tree threshold.

\subsection{Tree Construction and Direct Reconstruction}
\label{sec:appendix-construction}

\paragraph{Tree construction.}
RGB values $I\in[0,1]^3$ are standardized as
$V=(I-\boldsymbol{\mu})/\boldsymbol{\sigma}$, with
$\boldsymbol{\mu}=(0.485,0.456,0.406)$ and
$\boldsymbol{\sigma}=(0.229,0.224,0.225)$.
Descriptor fitting and maximum-error subdivision tests use this standardized
space; reconstruction metrics use RGB values in $[0,1]$.
Starting from the depth-3 partition, we fit RGB values and gradients within
each of a candidate cell's 32 subregions using \cref{eq:local-ls}, then evaluate
its maximum approximation error. A cell is retained as a leaf if its error
does not exceed the threshold; otherwise, it is split into eight children,
which are tested recursively. The thresholds at depths 3, 4, and 5 are
0.7, 0.8, and 1.0, respectively; depth-6 cells are retained without further
subdivision. The same schedule is used for reconstruction and recognition.
At depth 6, each leaf still stores a $2\times4\times4$ lattice, but each
subregion contains one voxel. Its gradient channels use spatial and temporal
finite differences from the original video neighborhood. These channels
provide encoder features; the singleton subregion's zero local coordinates
make its analytic reconstruction depend only on the stored RGB value.
The leaves cover the full video without overlap, and their geometric centers
determine the Morton ordering used for subsequent sequence processing.
\Cref{fig:direct-threshold-sweep} shows how scaling these thresholds trades
the number of leaves for direct reconstruction quality.
Under the VAE protocol, the mean leaf counts at depths 3--6 are
28.06, 413.2, 3,073, and 143.4, respectively. These account for 0.8\%, 11.3\%,
84.0\%, and 3.9\% of the 3,657.66 mean leaves.

\paragraph{Efficient implementation.}
We fit and test candidate cells across depths in parallel on the GPU, then
select leaves using ancestor split masks and pack them in Morton order.
Convolutional residual features are precomputed concurrently with construction.
This highly parallelized implementation takes under 2 ms for construction and
approximately 22 ms for construction, encoding, and decoding per video on an
RTX 4090, avoiding per-cell Python recursion and pointer-based traversal.

\paragraph{Direct reconstruction.}
\label{sec:appendix-direct-reconstruction}
At each video coordinate, we find the covering leaf $c$ and its subregion
$q$, then evaluate $\hat V_{c,q}$. Combining these local predictions recovers
the complete video without a VAE or learned residual features. With PSNR
computed per clip and then averaged, this analytic reconstruction achieves
34.72 dB on K400. 

\Cref{fig:k400-direct-reconstruction} compares
analytic reconstructions with controlled variations in cell content and placement.

\subsection{Video Generation Protocol}
\label{sec:appendix-generation}

\paragraph{Data and preprocessing.}
We train the class-conditional generator on all 101 UCF101 classes using
9,537 training videos. The 3,783 test videos provide the real reference for
evaluating the final 150,000-step EMA checkpoint.
We follow the split-1 source-group partition: g01--g07 for testing and
g08--g25 for training.

Each effective clip contains 29 frames at 12 fps. For source frame rate $f$,
the sampled indices are $\operatorname{round}(a+jf/12)$ for $j=0,\ldots,28$,
clamped to the last available frame. Images are aspect-ratio-preserving
bilinearly resized to a short edge of 256 with antialiasing, then center-cropped
to $256\times256$. No online spatial crop, flip, or color augmentation is used.
Each test video contributes one temporal-center clip.

\paragraph{Latent extraction.}
The frozen K400-trained VAE receives 32 frames by repeating the last valid
frame three times, with the same ImageNet RGB normalization as in VAE
reconstruction. We use posterior means with latent shape
$(T,C,H,W)=(8,16,32,32)$ and standardize each channel using means and
standard deviations estimated from training latents.

\paragraph{Generator and optimization.}
The Latte-XL/2-style backbone has 28 alternating spatial/temporal blocks,
width 1,152, 16 attention heads, MLP ratio 4, and $1\times2\times2$ latent
patches, yielding 2,048 tokens. It has 674,195,392 total parameters and
673,891,264 trainable parameters. A single model is trained from random
initialization on all classes, without pretrained generation weights or LoRA.
Class and time embeddings condition AdaLN-Zero blocks; class dropout is 0.1.
There is no text encoder.

For normalized latent $x_0$ and Gaussian noise $\epsilon$, we sample
$u\sim\mathcal{N}(0,1)$ and set $\sigma=\operatorname{sigmoid}(u)$.
The model receives $x_\sigma=(1-\sigma)x_0+\sigma\epsilon$ and time
$1000\sigma$, and minimizes unweighted velocity MSE against $\epsilon-x_0$.
Training uses AdamW with constant learning rate $10^{-4}$, no warm-up,
$\beta=(0.9,0.999)$, zero weight decay, gradient-norm clipping at 1.0,
and EMA decay 0.999. We use eight GPUs with five samples per GPU and no
gradient accumulation, for a global batch of 40 over 150,000 updates.
Computation uses BF16 autocast with FP32 parameters and optimizer states.
The training seed is 2026. Training samples are shuffled without
class-balanced training sampling.

\paragraph{Sampling and evaluation.}
Sampling uses the 150,000-step EMA checkpoint, standard Gaussian initialization,
50 Euler updates on a uniform noise schedule from 1 to 0, and classifier-free
guidance of 4.0 on all 16 velocity channels. After inverse latent normalization,
the fixed VAE decodes without an input video, reference tree, or teacher split
mask. Outputs are cropped to the first 29 frames and clamped to RGB $[0,1]$.
We report FVD 264 and KVD 28.6 from the same evaluation with 5,050 generated
clips and 3,783 original RGB test clips as the real reference.

Features are extracted directly from decoded RGB tensors, without MP4
compression, using the StyleGAN-V I3D TorchScript extractor with 400-dimensional
outputs. Inputs are rounded to uint8 RGB, with rescaling and resizing enabled.
FVD uses population covariances ($\mathrm{ddof}=0$). KVD is unbiased squared
MMD with $k(a,b)=(a^\top b/400+1)^3$.

\paragraph{Qualitative examples.}
\Cref{fig:latent-generation} shows selected excerpts with four frames spaced
by 0.25 s in each row, retaining their full decoded spatial extent.
Quantitative evaluation uses the generated set across all 101 classes.

\clearpage
\subsection{Kinetics-400 VAE Reconstructions}
\suppressfloats[t]
\begin{figure}[!htbp]
  \centering
  \includegraphics[width=\textwidth]{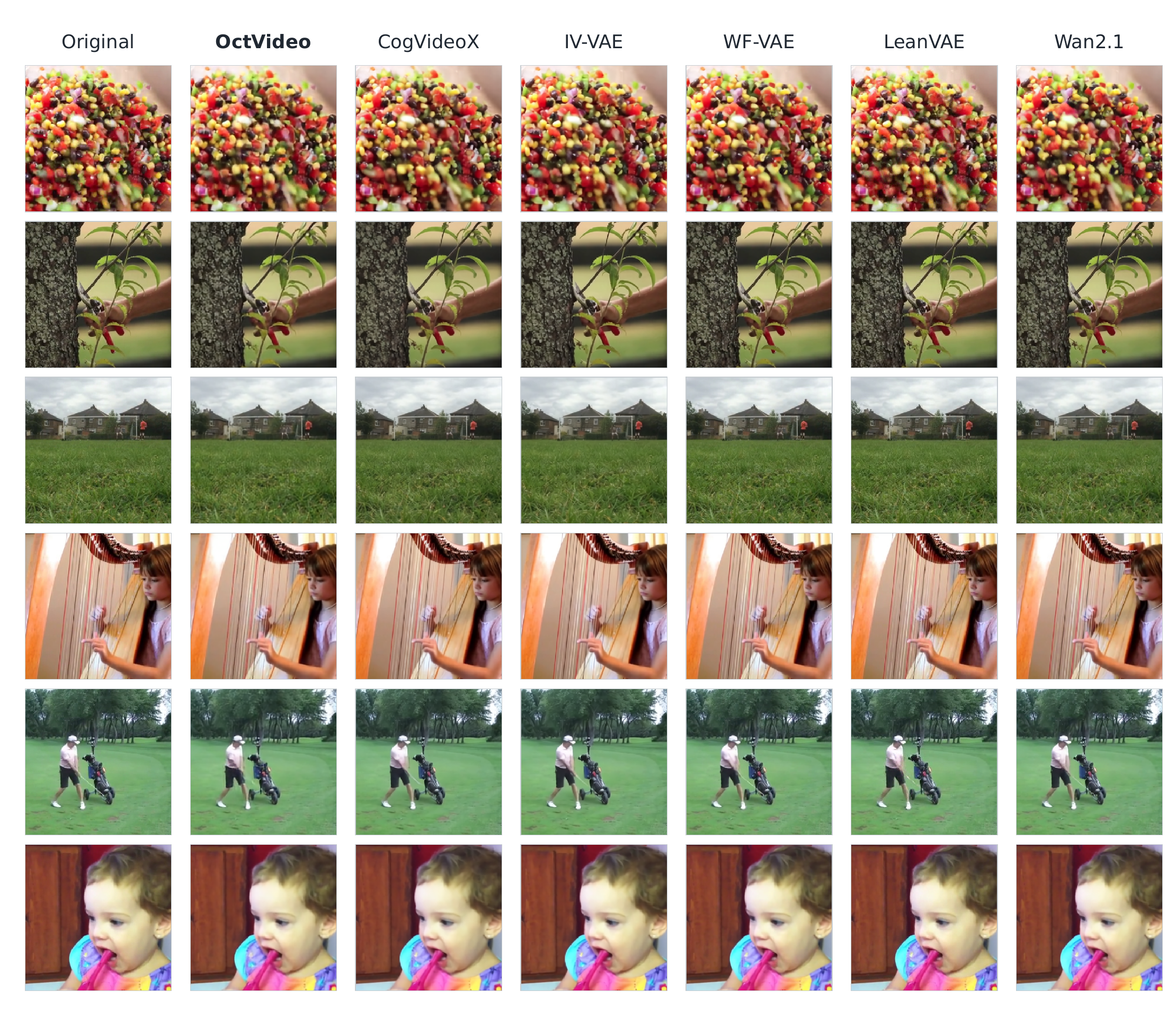}
  \caption{Qualitative VAE reconstructions on Kinetics-400. Each row shows
  the same source frame across the original clip and six reconstructions.}
  \Description{Six Kinetics-400 clips, each compared across the original
  frame and six video VAE reconstructions.}
  \label{fig:k400-vae-qualitative}
\end{figure}

\FloatBarrier
\end{document}